# A Novel Challenge Set for Hebrew Morphological Disambiguation and Diacritics Restoration

**Avi Shmidman[1,2], Joshua Guedalia[1,2], Shaltiel Shmidman[1,2], Moshe Koppel[1,2], Reut Tsarfaty[1,3]**
[1]Bar Ilan University / Ramat Gan, Israel   [2]DICTA / Jerusalem, Israel
[3]Allen Institute for Artificial Intelligence
{avi.shmidman,josh.guedalia,shaltiel.shmidman,
moshe.koppel,reut.tsarfaty}@biu.ac.il

## Abstract

One of the primary tasks of morphological parsers is the disambiguation of homographs. Particularly difficult are cases of unbalanced ambiguity, where one of the possible analyses is far more frequent than the others. In such cases, there may not exist sufficient examples of the minority analyses in order to properly evaluate performance, nor to train effective classifiers. In this paper we address the issue of unbalanced morphological ambiguities in Hebrew. We offer a challenge set for Hebrew homographs — the first of its kind — containing substantial attestation of each analysis of 21 Hebrew homographs. We show that the current SOTA of Hebrew disambiguation performs poorly on cases of unbalanced ambiguity. Leveraging our new dataset, we achieve a new state-of-the-art for all 21 words, improving the overall average F1 score from 0.67 to 0.95. Our resulting annotated datasets are made publicly available for further research.

## 1 Introduction

It is a known phenomenon that the distribution of linguistic units, or words, in a language follows a *Zipf law* distribution (Zipf, 1949), wherein a relatively small number of words appear frequently, and a much larger number of items appear in a long tail of words, as rare events (Czarnowska et al., 2019). Significantly, this also applies to the distribution of analyses of a given homograph. Take for instance the simple POS-tag ambiguity in English between noun and verb (Elkahky et al., 2018). The word "fair" can be used as an adjective ("a fair price") or as a noun ("she went to the fair"). Yet, the distribution of these two analyses is certainly not fair; the adjectival usage is far more frequent than the nominal usage (e.g., in Bird et al. (2008) the latter is six times more frequent than the former). We will call such cases "unbalanced homographs".

Cases of unbalanced homographs pose a formidable challenge for automated morphological parsers and segmenters. In tagged training corpora, the frequent option will naturally dominate the overwhelming majority of the occurrences. If the training corpus is not sufficiently large, then the sparsity of the minority analysis will prevent generalization by machine-learning models. By the same token, it can be difficult to evaluate the performance of tagging systems regarding unbalanced homographs, because the sparsity of the minority analysis prevents computation of adequate scoring.

The empirical consequences of unbalanced homographs are magnified in morphologically rich languages (MRLs), including many Semitic languages, where distinct morphemes are often affixed to the word itself, resulting in additional ambiguity (Fabri et al., 2014; Habash et al., 2009). Furthermore, in many Semitic MRLs, the letters are almost entirely consonantal, omitting vowels. This results in a particularly high number of homographs, each with a different pronunciation and meaning.

In this paper, we focus upon unbalanced homographs in Hebrew, a highly ambiguous MRL in which vowels are generally omitted (Itai and Wintner, 2008; Adler and Elhadad, 2006). Take for example the Hebrew word מדינה. This frequent word is generally read as a single nominal morpheme, מְדִינָה, meaning "country". However, it can also be read as מִדִּינָה, "from the law/judgment of her", wherein the initial and final letters both serve as distinct morphemes. This last usage is far less common, and, in an overall distribution, it would be relegated to the long tail, with very few attestations in any given corpus.

Hebrew is a low resource language, and as such, the problem of unbalanced homographs is particularly acute. Existing tagged corpora of Hebrew are of limited size, and in most cases of unbalanced homographs, the corpora do not provide sufficient examples to evaluate performance regarding minority analyses, nor to train an effective classifier.

Here, we propose to overcome this difficulty by means of a challenge set: a group of specialized training sets which each focus upon one particular homograph, offering substantial attestations of the competing analysis. Designing such *contrast sets* that expose particularly hard unbalanced cases was recently proposed as a complementary evaluation effort for a range of NLP tasks by Gardner et al. (2020). Notably, all tasks therein focus exclusively on English, and do not make any reference to morphology. Another, particularly successful, instance of this approach is the Noun/Verb challenge set for English built by Elkahky et al. (2018). Yet, heretofore, no challenge sets have been built to address cases of unbalanced homographs in Hebrew.

In order to fill this lacuna, we built a challenge set for 12 frequent cases of unbalanced Hebrew homographs. Each of these words admits of two possible analyses, each with its own diacritization and interpretation.[1] For each of the possible analyses, we gather 400 2,500 sentences exemplifying such usage, from a varied corpus consisting of news, books, and Wikipedia. Furthermore, in order to highlight the particular problem regarding unbalanced homographs, we add an additional 9 cases of balanced homographs, for contrast and comparison. All in all, the corpus contains over 56K sentences.[2]

## 2 Description of the Corpus

In Table 1 we list the 21 homographs addressed in our challenge set. For each case, we specify the frequency of each analysis in naturally-occurring Hebrew text, and the ratio between them.[3] The 21 homographs include a wide range of homograph types. Some are cases of different POS types: Adj vs. Prep (13), Noun vs. Verb (15, 18), Pronoun vs. Prep (2,4), Noun vs. Prep (9), etc. Other cases differ in terms of whether the final letter should be segmented as a suffix (10, 13, 20). In some instances, the morphology is the same, but the difference lies in the stem/lexeme (5, 7, 8, 11).

In choosing our 21 homographs, we first assembled a list of the most frequent homographs in the Hebrew language. For the simplicity of this initial proof of concept, we constrained our list to homographs with only two primary analyses. We also constrained our list to cases where the two analyses represent different lexemes, skipping over cases in which the difference is only one of inflection. Further, some cases were filtered out due to data sparsity. Finally, we also included a number of less frequent homographs, to allow for a comparison between frequent and infrequent homographs.

In order to gather sentences for the contrast sets, we first sampled 5000 sentences for each target word, and sent them to student taggers. For balanced homographs, with ratios of 1:3 or less, this process handily provided a sufficiently large number of sentences for each of the two analyses. However, regarding cases of unbalanced homographs, wherein the naturally occurring ratio of the minority analysis can be 30:1 or even 129:1, this initial corpus was far from adequate. We used two methods to identify additional candidate sentences: (1) We ran texts through an automated Hebrew diacritizer (Shmidman et al., 2020) and took the cases where the word was diacritized as the minority analysis. (2) Where relevant, we leveraged optional Hebrew orthographic variations which indicate that a given word is intended in one specific way. These candidate sentences were then sent to student taggers to confirm that the minority analysis was in fact intended. Our student taggers tagged approximately 300 sentences per hour. Evaluation of their work revealed that they averaged an accuracy of 98 percent. In order to overcome this margin of error, we employed a Hebrew language expert who proofread the resulting contrast sets. In our final corpus, each analysis of each homograph is attested in at least 400 sentences, and usually in 800-2.5K sentences (full details in Appendix Table 1).

One issue we encountered when collecting naturally-occurring Hebrew sentences is that a small number of specific word-neighbors and collocations tend to dominate the examples. As an example: the word אפשר can be vocalized as אֶפְשָׁר ("possible", the majority case), or אִפְּשֵׁר ("he allowed"). However, over one third of the naturally occurring cases of the majority case boil down to some 90 frequently occurring collocations, such as אי אפשר ("impossible") or האם אפשר ("is it possible?"). As such, a machine-learning model would overfit to those specific collocations, rather than learning more generic overarching patterns of the word usage. Therefore, we constrained our data collection such that there may be no more than 20 cases of any given word neighbor combination.[4]

---
[1]In some of the cases, additional analyses are theoretically possible, but exceedingly rare.
[2]In cases where a given sentence contains more than one instance of the target word, the sentence is included multiple times, once for each instance.
[3]All statistics in this paper regarding the distribution of Hebrew word analyses are based upon an in-house annotated 2.4M word corpus maintained by DICTA.

Table 1: The homographs covered in our challenge set. Words 1-12 are unbalanced homographs, in which the ratio between the two analyses is particularly skewed. These cases pose a particularly difficult disambiguation challenge because they are severely underrepresented in existing tagged Hebrew corpora.

| # | Form | Option 1 | | | Option 2 | | | Ratio |
|---|---|---|---|---|---|---|---|---|
| | | Word (Translation) | Morphology | Count / 1M | Word (Translation) | Morphology | Count / 1M | |
| 1 | את | אֵת ([accusative]) | ACC | 18164 | אַתְּ (you) | Pronoun [F,S,2] | 275 | 66:1 |
| 2 | אתה | אַתָּה (you) | Pronoun [M,S,2] | 1430 | אִתָּהּ (with her) | Prep+Suf_Pron [F,S,3] | 26 | 55:1 |
| 3 | אתכם | אֶתְכֶם (you) | ACC+Suf_Pron [M,P,2] | 70 | אִתְּכֶם (with you) | Prep+Suf_Pron [M,P,2] | 7 | 10:1 |
| 4 | אתם | אַתֶּם (you) | Pronoun [M,P,2] | 324 | אִתָּם (with them) | Prep+Suf_Pron [M,P,3] | 34 | 10:1 |
| 5 | ברכת | בִּרְכַּת (blessing) | Noun [cons,F,S] | 25 | בְּרֵכַת (pool) | Noun [cons,F,S] | 0.8 | 30:1 |
| 6 | הרי | הֲרֵי (indeed) | Conj / Intj | 418 | הָרֵי (mountains) | Noun [cons,M,P] | 12 | 33:1 |
| 7 | יאמר | יֹאמַר (he will say) | Verb [M,S,3,FUTURE] | 18 | יֵאָמֵר (will be said) | Verb [M,S,3,FUTURE] | 0.4 | 43:1 |
| 8 | מסכת | מַסֶּכֶת (tractate) | Noun [abs/cons,F,S] | 54 | מַסֵּכָה (mask) | Noun [cons,F,S] | 1 | 43:1 |
| 9 | עם | עִם (with) | Preposition | 4240 | עַם (nation) | Noun [abs/cons,M,S] | 286 | 14:1 |
| 10 | פניה | פָּנֶיהָ (her face) | Noun [F,M,P,suf=F,S,3] | 55 | פְּנִיָּה (application) | Noun [F,S] | 2 | 33:1 |
| 11 | פרשו | פָּרְשׁוּ (they left) | Verb [MF,P,3,PAST] | 6 | פֵּרְשׁוּ (they interpreted) | Verb [MF,P,3,PAST] | 0.4 | 15:1 |
| 12 | שלישית | שְׁלִישִׁית (third) | Cardinal [F,S] | 107 | שְׁלִישִׁיָּה (trio) | Noun [cons,F,S] | 0.8 | 129:1 |
| 13 | אחר | אַחֵר (different) | Adj [M,S] | 474 | אַחַר (after) | Preposition | 387 | 1:1 |
| 14 | בניה | בָּנֶיהָ (her sons) | Noun [M,P,suf=F,S,3] | 8 | בְּנִיָּה (building) | Noun [F,S] | 5 | 1 5:1 |
| 15 | חזרה | חֲזָרָה (returning) | Noun [F,S] | 62 | חָזְרָה (she returned) | Verb [F,S,3,PAST] | 55 | 1:1 |
| 16 | ידע | יָדַע (he knew) | Verb [M,S,3,PAST] | 88 | יֶדַע (knowledge) | Noun [abs/cons,M,S] | 55 | 1.5:1 |
| 17 | כשר | כְּשַׂר (as minister) | Prep+Noun [abs/cons,M,S] | 35 | כָּשֵׁר (kosher) | Adj [M,S] / Propn [MF,S] | 14 | 2.5:1 |
| 18 | כתב | כָּתַב (he wrote) | Verb [M,S,3,PAST] | 252 | כְּתָב (writing) | Noun [cons,M,S] | 103 | 2.5:1 |
| 19 | מבין | מֵבִין (understands) | Participle [M,S] | 174 | מִבֵּין (from amongst) | Preposition | 98 | 2:1 |
| 20 | ספריה | סְפָרֶיהָ (her books) | Noun [M,P,suf=F,S,3] | 13 | סִפְרִיָּה (library) | Noun [F,S] | 4 | 2.5:1 |
| 21 | עמנו | עַמֵּנוּ (our nation) | Noun [M,S,suf=MF,P,1] | 23 | עִמָּנוּ (with us) | Prep+Suf_Pron [MF,P,1] | 12 | 2:1 |

## 3 Experiments

We first use our challenge set to evaluate current state-of-the-art performance on the morphological disambiguation of Hebrew homographs. The best existing tool for Hebrew morphological disambiguation is YAP: *Yet Another Parser* (Tsarfaty et al., 2019). We run all 56,000+ sentences from our challenge set through YAP. Due to the unbalanced natural distribution of the possible analyses in many of the cases, we compute recall and precision results separately for each analysis, and we then compute a macro-averaged F1 score.

Next, we use our challenge set to train classifiers for each of the homographs in our corpus. We implement 2 layer MLPs using the DyNet framework (Neubig et al., 2017). As input, we feed the MLP an encoding $h(w_i)$, a representation of the context of the target word within the sentence. The target word itself is masked and not included in the input. The output of the MLP is a probabilistic choice of either Class 1 or Class 2, where each class represents one of the two possible diacritization options.

We applied two methods to represent the surrounding context in the MLP input. The first is encoding the three neighboring words on both sides of the target word;[5] see Equation 1. The second is encoding the whole sentence around the word using a 2 layer biLSTM (Hochreiter and Schmidhuber, 1997), Equation 2.

| | | YAP | | | | |
|---|---|---|---|---|---|---|
| | | Option 1 | | Option 2 | | |
| # | Word | Prec | Recall | Prec | Recall | Avg-F1 |
| 1 | את | 85.61 | 99.24 | 100.00 | 12.37 | .570 |
| 2 | אתה | 53.55 | 96.42 | 95.04 | 21.48 | .519 |
| 3 | אתכם | 69.30 | 97.26 | 71.88 | 13.71 | .520 |
| 4 | אתם | 37.87 | 99.87 | 75.00 | .24 | .277 |
| 5 | ברכת | | .00 | 58.31 | 93.20 | — |
| 6 | הרי | 92.53 | 97.10 | 88.82 | 63.04 | .843 |
| 7 | יאמר | | 00 | 52 19 | 100.00 | — |
| 8 | מסכת | 86.93 | 24.84 | 41.51 | 89.86 | .477 |
| 9 | עם | 87 73 | 99 20 | 91 59 | 36.03 | 724 |
| 10 | פניה | 28.36 | 33.98 | 82.90 | 78.85 | .559 |
| 11 | פרשו | 71 93 | 90.82 | | 00 | — |
| 12 | שלישית | 75.12 | 90.60 | 93.38 | 65.13 | .794 |
| 13 | אחר | 95 73 | 88.84 | 82 79 | 90.66 | .894 |
| 14 | בניה | 45.22 | 27.29 | 84.67 | 85.51 | .596 |
| 15 | חזרה | 81.03 | 66.49 | 76.84 | 87.64 | .775 |
| 16 | ידע | 85.09 | 63.50 | 95.76 | 89.63 | .827 |
| 17 | כשר | 94.79 | 63.13 | 75.11 | 66.45 | .732 |
| 18 | כתב | 97.63 | 78.17 | 72.61 | 90.86 | .838 |
| 19 | מבין | 77.03 | 86.32 | 94.84 | 90.48 | .870 |
| 20 | ספריה | 87.93 | 14.98 | 75.25 | 99.15 | .556 |
| 21 | עמנו | 83.76 | 38.89 | 76.65 | 96.38 | .693 |

Table 2: Results running our entire challenge set through YAP, the SOTA Hebrew morphological tagger. YAP performs far better on the balanced cases (13 21) than on the unbalanced cases (1 12). It is also worth noting that the YAP's poor performance on unbalanced homographs is not tied to the overall frequency of the word; the particularly frequent words (1,2,4,6,9) demonstrate similar scores to those of the relatively infrequent words (8,10,12). In three cases (5,7,11), where the difference is only the lexeme/stem, YAP always chooses one option; hence the − scores.

---

[4]Our challenge set is available for use in future research.

[5]The efficacy of Hebrew homograph disambiguation via

| # | Word | Word2vec | | Morphology | | Composite | |
|---|------|----------|------|------------|------|-----------|------|
|   |      | Concat | LSTM | Concat | LSTM | Concat | LSTM |
| 1 | את | .955 | .953 | .946 | .940 | **.969** | .958 |
| 2 | אתה | .945 | .963 | .909 | .934 | .958 | **.967** |
| 3 | אתכם | .915 | .919 | .814 | .831 | **.922** | .940 |
| 4 | אתם | .941 | .953 | .924 | .933 | .944 | **.959** |
| 5 | ברכת | .951 | **.968** | .733 | .805 | .936 | .965 |
| 6 | הרי | .960 | .966 | .923 | .931 | **.974** | .969 |
| 7 | יאמר | .859 | **.893** | 805 | 851 | 878 | .885 |
| 8 | מסכת | 950 | **.972** | 849 | 869 | 954 | 966 |
| 9 | עם | .894 | **.917** | 838 | 850 | 891 | 911 |
| 10 | פניה | 930 | 942 | 870 | 893 | 943 | **.946** |
| 11 | פרשו | 935 | 957 | 881 | 916 | 948 | **.963** |
| 12 | שלישית | .953 | **.969** | .899 | .922 | .955 | .966 |
| 13 | אחר | 965 | **.976** | 939 | 935 | 969 | **.976** |
| 14 | בניה | 952 | **.965** | 855 | 883 | 947 | 964 |
| 15 | חזרה | .925 | **.951** | .861 | .893 | .935 | .949 |
| 16 | ידע | .957 | .955 | .910 | .907 | .963 | **.966** |
| 17 | כשר | .953 | **.974** | .889 | .912 | .964 | .971 |
| 18 | כתב | .976 | .982 | .910 | .924 | .972 | **.983** |
| 19 | מבין | .976 | .975 | .966 | .970 | .976 | **.980** |
| 20 | ספריה | .930 | .945 | .856 | .875 | .938 | **.949** |
| 21 | עמנו | .920 | .915 | .888 | .872 | .923 | **.926** |

Table 3: Accuracy of our specialized classifiers for the 21 homographs in our challenge set. We evaluate three methods for encoding the context words, and we run each method two ways: (1) "Concat": concatenate encodings of 3 neighboring words on each side; (2) "LSTM": run complete sentence context through a BiLSTM. We show F1 scores for each, macro-averaged across the two classes. See Appendix Tables 4-5 for a breakdown of recall/precision scores for each analysis.

| | | Unbalanced | | | | Balanced | |
|---|------|------|------|---|------|------|------|
| # | Word | YAP | Ours | # | Word | YAP | Ours |
| 1 | את | .570 | **.969** | 13 | אחר | .894 | **.969** |
| 2 | אתה | .519 | **.958** | 14 | בניה | .596 | **.947** |
| 3 | אתכם | .520 | **.922** | 15 | חזרה | .775 | **.935** |
| 4 | אתם | .277 | **.944** | 16 | ידע | .827 | **.963** |
| 5 | ברכת |  | **.936** | 17 | כשר | 732 | **.964** |
| 6 | הרי | 843 | **.974** | 18 | כתב | 838 | **.972** |
| 7 | יאמר | – | **.878** | 19 | מבין | .870 | **.976** |
| 8 | מסכת | .477 | **.954** | 20 | ספריה | .556 | **.938** |
| 9 | עם | .724 | **.891** | 21 | עמנו | .693 | **.923** |
| 10 | פניה | .559 | **.943** | | | | |
| 11 | פרשו |  | **.948** | | | | |
| 12 | שלישית | 794 | **.955** | | | | |

Table 4: Comparison of the SOTA morphological dis ambiguation of Hebrew homographs (YAP) to our spe cialized classifiers (Avg F1). See Appendix Table 3 for a full precision/recall breakdown of this comparison.

(1) $h(w_i) = w_{i-3} \cdot w_{i-2} \cdot w_{i-1} \cdot w_{i+1} \cdot w_{i+2} \cdot w_{i+3}$

(2) $h(w_i) = LSTM(w_{0:i}) \cdot LSTM(w_{n:i})$

We explore three alternate methods of encoding the vector $w_i$. Our initial approach uses pre-trained word2vec embeddings for the neighboring words.[6]

Our second approach uses morphological information about the context words. Of course, we don't have any a priori knowledge regarding the morphological tagging of the neighboring words; and indeed, in a large percentage of the cases, the morphology of the neighboring words can be resolved in multiple ways. Thus, we construct a lattice of all possible analyses of the context words. For every context word $w_i$, we construct a vector for each possible part of speech $pos_j$ containing a trainable embedding for each possible morphological feature. The vector thus encodes: part-of-speech, gender, number, person, status, binyan, suffix, suf_gender, suf_person, suf_number, prefix.[7] If a feature is not applicable to $w_i$, we simply assign an NA embedding. We concatenate each vector $w^i_{pos_j}$ into a single vector representing $w_i$.

Finally, we explore a third composite method in which we concatenate the encodings from the two previous methods to the encoding for $w_i$.

We run each contrast set using each of our three methods for encoding the neighboring words. We evaluate the results using 10-fold cross validation.

## 4 Results and Analysis

In Table 2, we display the results of our baseline experiment, where we evaluate current SOTA (YAP) performance on our challenge set. These results empirically demonstrate how much more difficult it is for YAP to resolve the cases of unbalanced homographs. The unbalanced cases are shown in the top half of the table (1 12). YAP's F1 score is below .8 for all but one of the cases, and it is below .6 for 9 out of the 12 cases. In the two cases of Pronoun vs. Suffixed Preposition (2,4), YAP performs particularly poorly, scoring .4 and .1. In contrast, the bottom half of the table (13-21) details

---

short contexts was demonstrated by Fraenkel et al (1979); Choueka and Lusignan (1985). Regarding short-context disambiguation methods in general, see Hearst (1991); Yarowsky (1994).

[6]We use word2vecf (Levy and Goldberg, 2014) to build syntax-sensitive word embeddings, based on a corpus of 400M words of Hebrew text. To be sure, BERT might seem the more obvious choice rather than word2vec. However, BERT has been shown to be somewhat ineffective for morphologically rich languages such as Hebrew (Tsarfaty et al., 2020). BERT-based models underperform YAP and perform at the same level as BILSTM-based models, and BERT fails to capture internal morphological complexity (Klein and Tsarfaty, 2020).

[7]For verbs only, we add a morphosyntactic valence feature indicating the transitivity of the general usage of the verb. This is reminiscent of supertagging (Bangalore and Joshi, 1999) and shows non-negligible empirical contribution on our data. See Appendix Table 2 for a comparison of results with and without the valence feature.

nine cases of balanced homographs. As expected, YAP does considerably better here: all F1 scores are above .5, and four of the cases are above .8. The weakest cases are those in which YAP has to differentiate between an unsegmented noun and a case of a noun plus possessive suffix (cases 14,20). In both of these cases, YAP scores an F1 of approximately .56 (which, interestingly, is precisely on par with the analogous unbalanced case [10]).

In Table 3, we display results regarding our specialized classifiers. In most cases, using a biLSTM over the entire sentence context performs better than a concatenation of the three neighbor words on each side. In terms of the encoding method for the context words, word2vec performs better than the morphological lattice. This may be because word2vec can better represent the regularly expected usage of the neighboring words, while the morphology lattice represents all possible analyses with equal likelihood. A second possibility is that the contrast sets were not sufficiently large to optimally train the embeddings of the morphological characteristics, whereas word2vec embeddings have the benefit of pretraining on over 100M words. The combination of the latter two methods overall outperforms each one of them individually; thus, although word2vec succeeds in encoding most of what is needed to differentiate between the options, the information provided by the morph lattice sometimes helps to make the correct call.

In Table 4, we compare the results of our composite-method with those of YAP. Our specialized classifiers set a new SOTA for all the cases.

## 5 Related Work

Many recent papers have proposed global or unsupervised methods for homograph disambiguation in English (e.g. Liu et al. (2018); Wilks and Stevenson (1997); Chen et al. (2009)). While such methods have obvious advantages, they have limited applicability to Hebrew. As noted, in Hebrew the majority of the words are ambiguous, including the core building blocks of the language; without these anchors, global approaches tend to result in poor performance regarding unbalanced homographs.

The problem of Hebrew diacritization is analogous to that of Arabic diacritization; Arabic, like Hebrew, is a morphologically-rich language written without diacritics, resulting in high ambiguity. Many recent studies have proposed machine-learning approaches for the prediction of Arabic diacritics across a given text (e.g. Bebah et al. (2014); Belinkov and Glass (2015); Neme and Paumier (2019); Fadel et al. (2019a,b); Darwish et al. (2020). However, these studies all perform evaluations on standard Arabic textual datasets, and do not evaluate accuracy regarding minority options of unbalanced homographs. We believe that these models would likely benefit from specialized challenge sets of the sort presented here to overcome the specific hurdle of unbalanced homographs.

## 6 Conclusion

Due to high morphological ambiguity, as well as the lack of diacritics, Semitic languages pose a particularly difficult disambiguation task, especially when it comes to unbalanced homographs. For such cases, specialized contrast sets are needed, both in order to evaluate performance of existing tools, as well as in order to train effective classifiers. In this paper, we construct a new challenge set for Hebrew disambiguation, offering comprehensive contrast sets for 21 frequent Hebrew homographs. These contrast sets empirically demonstrate the limitations of reported SOTA results when it comes to unbalanced homographs; a model may report a SOTA for a benchmark, yet fail miserably on real world rare-but important cases. Our new corpus will allow Hebrew NLP researchers to test their models in an entirely new fashion, evaluating the ability of the models to predict minority homograph analyses, as opposed to existing Hebrew benchmarks which tend to represent the language in terms of its majority usage. Furthermore, our corpus will allow researchers to train their own classifiers and leverage them within a pipeline architecture. We envision the classifiers positioned at the beginning of the pipeline, disambiguating frequent forms from the get go, and yielding improvement down the line, ultimately improving results for downstream tasks (e.g. NMT). Indeed, as we have demonstrated, neural classifiers trained on our contrast sets handily achieve a new SOTA for all of the homographs in the corpus.

## 7 Acknowledgements

The work of the last author has been supported by an ERC StG grant #677352 and an ISF grant #1739/26. We acknowledge the substantial help of our programmers, Yehuda Broderick and Cheyn Shmuel Shmidman.

Appendix

| # | Form | Option 1 Word (Translation) | Option 1 Morphology | Option 1 # sentences | Option 2 Word (Translation) | Option 2 Morphology | Option 2 # sentences |
|---|---|---|---|---|---|---|---|
| 1 | את | אֵת ([accusative]) | ACC | 2,402 | אַתְּ (you) | Pronoun [F,S,2] | 443 |
| 2 | אתה | אַתָּה (you) | Pronoun [M,S,2] | 2,198 | אִתָּהּ (with her) | Prep+Suf_Pron [F,S,3] | 2,450 |
| 3 | אתכם | אֶתְכֶם (you) | ACC+Suf_Pron [M,P,2] | 1,630 | אִתְּכֶם (with you) | Prep+Suf_Pron [M,P,2] | 816 |
| 4 | אתם | אַתֶּם (you) | Pronoun [M,P,2] | 1,474 | אִתָּם (with them) | Prep+Suf_Pron [M,P,3] | 2,064 |
| 5 | ברכת | בִּרְכַּת (blessing) | Noun [cons,F,S] | 1,027 | בְּרֵכַת (pool) | Noun [cons,F,S] | 1,384 |
| 6 | הרי | הֲרֵי (indeed) | Conj / Intj | 1,939 | הָרֵי (mountains) | Noun [cons,M,P] | 419 |
| 7 | יאמר | יֹאמַר (he will say) | Verb [M,S,3,FUTURE] | 838 | יֵאָמֵר (will be said) | Verb [M,S,3,FUTURE] | 922 |
| 8 | מסכת | מַסֶּכֶת (tractate) | Noun [abs/cons,F,S] | 975 | מַסֵּכָת (mask) | Noun [cons,F,S] | 562 |
| 9 | עם | עִם (with) | Preposition | 2,416 | עַם (nation) | Noun [abs/cons,M,S] | 510 |
| 10 | פניה | פָּנֶיהָ (her face) | Noun [F,M,P,suf=F,S,3] | 607 | פְּנִיָּה (application) | Noun [F,S] | 2,435 |
| 11 | פרשו | פָּרְשׁוּ (they left) | Verb [MF,P,3,PAST] | 1,321 | פֵּרְשׁוּ (they interpreted) | Verb [MF,P,3,PAST] | 482 |
| 12 | שלישית | שְׁלִישִׁית (third) | Cardinal [F,S] | 1,199 | שְׁלִישִׁיָּת (trio) | Noun [cons,F,S] | 1,285 |
| 13 | אחר | אַחֵר (different) | Adj [M,S] | 2,422 | אַחַר (after) | Preposition | 1,215 |
| 14 | בניה | בָּנֶיהָ (her sons) | Noun [M,P,suf=F,S,3] | 578 | בְּנִיָּה (building) | Noun [F,S] | 2,448 |
| 15 | חזרה | חֲזָרָה (returning) | Noun [F,S] | 960 | חָזְרָה (she returned) | Verb [F,S,3,PAST] | 1,212 |
| 16 | ידע | יָדַע (he knew) | Verb [M,S,3,PAST] | 651 | יֶדַע (knowledge) | Noun [abs/cons,M,S] | 1,538 |
| 17 | כשר | כְּשַׂר (as minister) | Prep+Noun [abs/cons,M,S] | 959 | כָּשֵׁר (kosher) | Adj [M,S] / Propn [MF,S] | 753 |
| 18 | כתב | כָּתַב (he wrote) | Verb [M,S,3,PAST] | 2,078 | כְּתָב (writing) | Noun [cons,M,S] | 721 |
| 19 | מבין | מֵבִין (understands) | Participle [M,S] | 891 | מִבֵּין (from amongst) | Preposition | 2,473 |
| 20 | ספריה | סְפָרֶיהָ (her books) | Noun [M,P,suf=F,S,3] | 664 | סִפְרִיָּה (library) | Noun [F,S] | 1,715 |
| 21 | עמנו | עַמֵּנוּ (our nation) | Noun [M,S,suf=MF,P,1] | 471 | עִמָּנוּ (with us) | Prep+Suf_Pron [MF,P,1] | 1,007 |

Table 1: The homographs covered in our challenge set, the possible analyses for each homograph, and the number of attestations in our challenge set of each homograph analysis.

| | | Composite Without Valence | | | | | Composite With Valence | | | | |
|---|---|---|---|---|---|---|---|---|---|---|---|
| | | Option 1 | | Option 2 | | | Option 1 | | Option 2 | | |
| # | Word | Prec | Recall | Prec | Recall | Avg-F1 | Prec | Recall | Prec | Recall | Avg-F1 |
| 1 | את | 98.33 | 99.24 | 95.81 | 91.18 | .961 | **98.69** | **99.36** | **96.51** | **93.07** | **.969** |
| 2 | אתה | 95.56 | **95.44** | 95.72 | 95.83 | .956 | **96.01** | 95.35 | 95.66 | **96.27** | **.958** |
| 3 | אתכם | 93.88 | 95.28 | 90.25 | 87.54 | .917 | **94.39** | **95.34** | **90.46** | **88.62** | **.922** |
| 4 | אתם | 93.47 | **93.23** | **95.88** | 96.04 | **.947** | 93.66 | 92.24 | 95.32 | **96.20** | .944 |
| 5 | ברכת | 92.67 | **91.64** | 93.73 | 94.52 | .931 | **93.72** | 91.54 | 93.72 | **95.37** | **.936** |
| 6 | הרי | 98.70 | 98.70 | 94.10 | 94.10 | .964 | **99.00** | **99.10** | **95.90** | **95.46** | **.974** |
| 7 | יאמר | 86.70 | 86.70 | 87.75 | 87.75 | .872 | **87.60** | **86.81** | **87.95** | **88.68** | **.878** |
| 8 | מסכת | 96.46 | **96.91** | **94.27** | 93.46 | .953 | **96.99** | 96.45 | 93.53 | **94.49** | **.954** |
| 9 | עם | **95.40** | **98.08** | **89.85** | **78.27** | **.902** | 95.30 | 97.36 | 86.50 | 77.90 | .891 |
| 10 | פניה | 92.23 | **88.78** | 97.26 | 98.16 | .941 | **93.76** | 87.97 | 97.08 | **98.56** | **.943** |
| 11 | פרשו | 95.99 | **98.43** | **95.43** | 88.87 | .946 | **96.26** | 98.28 | 95.06 | **89.68** | **.948** |
| 12 | שלישית | 94.89 | **95.82** | **96.10** | 95.22 | .955 | **96.16** | 94.35 | 94.86 | **96.51** | .955 |
| 13 | אחר | 97.18 | 98.04 | 96.05 | 94.37 | .964 | **97.39** | **98.44** | **96.84** | **94.77** | **.969** |
| 14 | בניה | 91.25 | **90.17** | 97.68 | 97.95 | .943 | **92.68** | 90.17 | **97.69** | **98.31** | **.947** |
| 15 | חזרה | **93.96** | 91.34 | 93.32 | **95.37** | .935 | 93.40 | **91.96** | **93.73** | 94.88 | .935 |
| 16 | ידע | 93.49 | 93.91 | 97.36 | 97.17 | .955 | **94.40** | **95.25** | **97.94** | **97.56** | **.963** |
| 17 | כשר | **97.42** | 96.53 | 95.70 | **96.80** | **.966** | 96.93 | **96.63** | **95.79** | 96.16 | .964 |
| 18 | כתב | **98.52** | **99.05** | **97.13** | **95.56** | **.976** | 98.51 | 98.65 | 95.95 | 95.56 | .972 |
| 19 | מבין | **96.53** | 96.63 | 98.76 | **98.72** | **.977** | 96.12 | **96.74** | **98.80** | 98.56 | .976 |
| 20 | ספריה | **91.65** | 90.44 | 96.35 | **96.84** | .938 | 90.67 | **91.47** | **96.71** | 96.38 | .938 |
| 21 | עמנו | 88.96 | **88.07** | 94.30 | 94.75 | .915 | **91.48** | 87.48 | 94.11 | **96.08** | **.923** |

Table 2: Quantification of the contribution of the valence "supertag". We examine results of our "Concat Composite" method, wherein we use the three neighboring words before and after the homograph, with each neighboring word represented by a concatenation of its word2vec embedding and a lattice of the morphological features of the possible analyses of the word. We indicate the change in results when adding the valence supertag to the lattice.

|   |   | YAP | | | | | Our Classifier (Composite BiLSTM Method) | | | | |
|   |   | Option 1 | | Option 2 | | | Option 1 | | Option 2 | | |
| # | Word | Prec | Recall | Prec | Recall | Avg-F1 | Prec | Recall | Prec | Recall | Avg-F1 |
|---|---|---|---|---|---|---|---|---|---|---|---|
| 1 | את | 85.61 | **99.24** | **100.00** | 12.37 | .570 | **98.29** | 99.08 | 94.96 | **90.97** | .958 |
| 2 | אתה | 53 55 | **96.42** | **95.04** | 21.48 | 519 | 95.65 | **97.61** | **97.71** | 95.83 | .967 |
| 3 | אתכם | 69.30 | **97.26** | **71.88** | 13.71 | .520 | 95.51 | **96.54** | **92.90** | 90.90 | .940 |
| 4 | אתם | 37.87 | **99.87** | **75.00** | .24 | .277 | 94.11 | 95.66 | **97.33** | **96.36** | .959 |
| 5 | ברכת | – | .00 | 58.31 | 93.20 | – | **96.09** | 95.91 | 96.91 | **97.05** | .965 |
| 6 | הרי | 92.53 | **97.10** | **88.82** | 63.04 | .843 | **99.00** | 98.75 | 94.39 | **95.46** | .969 |
| 7 | יאמר | – | .00 | 52.19 | 100.00 | – | 86.71 | 89.74 | 90.24 | 87.33 | .885 |
| 8 | מסכת | 86.93 | 24.84 | **41 51** | **89 86** | 477 | **97.48** | 97.75 | 95.85 | **95.35** | .966 |
| 9 | עם | 87 73 | **99.20** | **91.59** | 36.03 | 724 | **96.25** | 97 64 | 88 36 | **82.50** | .911 |
| 10 | פניה | 28.36 | 33.98 | **82.90** | **78.85** | .559 | 92.79 | 89.92 | **97.53** | **98.28** | .946 |
| 11 | פרשו | 71.93 | 90.82 | – | .00 | – | **97.41** | 98.65 | 96.23 | **92.91** | .963 |
| 12 | שלישית | 75.12 | **90.60** | **93.38** | 65.13 | .794 | **96.86** | 96.07 | 96.39 | **97.12** | .966 |
| 13 | אחר | **95.73** | 88.84 | 82.79 | **90.66** | .894 | **97.90** | 98.96 | **97.89** | 95.80 | .976 |
| 14 | בניה | 45.22 | 27.29 | **84.67** | **85.51** | .596 | 96.12 | 92.37 | **98.21** | **99.12** | .964 |
| 15 | חזרה | **81 03** | 66.49 | 76 84 | **87.64** | 775 | **95.74** | 92.68 | 94.37 | **96.75** | .949 |
| 16 | ידע | 85.09 | 63.50 | **95.76** | **89.63** | .827 | 95.38 | 95.10 | **97.88** | **98.01** | .966 |
| 17 | כשר | **94.79** | 63.13 | 75.11 | **66.45** | .732 | **98.54** | 96.32 | 95.52 | **98.21** | .971 |
| 18 | כתב | **97.63** | 78.17 | 72.61 | **90.86** | .838 | **99.23** | 99.10 | 97.32 | **97.71** | .983 |
| 19 | מבין | 77.03 | 86.32 | **94.84** | **90.48** | .870 | 96.77 | 97.50 | **99.08** | **98.80** | .980 |
| 20 | ספריה | **87.93** | 14.98 | 75.25 | **99.15** | .556 | 92.15 | 93.24 | **97.39** | **96.95** | .949 |
| 21 | עמנו | **83 76** | 38.89 | 76.65 | **96.38** | 693 | **90.71** | 89.26 | 94.88 | 95 61 | .926 |

Table 3: Expanded results comparing the performance of our specialized classifiers with that of the state-of-the-art Hebrew morphological tagger, YAP. Our classifiers set a new SOTA for all cases, both balanced and unbalanced, although the improvement is much more substantial regarding the unbalanced cases. (In three cases [5,7,11], where the difference is only one of lexeme or verbal stem, YAP always chooses one option; hence the scores for these cases).

|   |   | Word2vec embeddings | | | | | Morphological characteristics | | | | | Composite Method | | | | |
|   |   | Option 1 | | Option 2 | | | Option 1 | | Option 2 | | | Option 1 | | Option 2 | | |
| # | Word | Prec | Recall | Prec | Recall | Avg-F1 | Prec | Recall | Prec | Recall | Avg-F1 | Prec | Recall | Prec | Recall | Avg-F1 |
|---|---|---|---|---|---|---|---|---|---|---|---|---|---|---|---|---|
| 1 | את | 98.29 | 98.88 | 93.93 | 90.97 | .955 | 97.93 | 98.68 | 92.78 | 89.08 | .946 | **98.69** | **99.36** | **96.51** | **93.07** | .969 |
| 2 | אתה | 93.95 | 94.67 | 94.95 | 94.27 | .945 | 90.51 | 90.75 | 91.28 | 91.06 | .909 | **96.01** | **95.35** | **95.66** | **96.27** | .958 |
| 3 | אתכם | 94.22 | 94.45 | 88.81 | 88.38 | .915 | 87.21 | 88.36 | 76.01 | 74.01 | .814 | **94.39** | **95.34** | **90.46** | **88.62** | .922 |
| 4 | אתם | 93.50 | 91.78 | 95.05 | 96.12 | .941 | 91.53 | 89.48 | 93.68 | 94.96 | .924 | **93.66** | **92.24** | **95.32** | **96.20** | .944 |
| 5 | ברכת | **94.26** | **94.70** | **95.98** | **95.65** | .951 | 70.77 | 67.29 | 76.17 | 79.00 | .733 | 93.72 | 91.54 | 93.72 | 95.37 | .936 |
| 6 | הרי | 98.74 | 98.35 | 92.65 | 94.33 | .960 | 96.78 | 97.80 | 89.52 | 85.26 | .923 | **99.00** | **99.10** | **95.90** | **95.46** | .974 |
| 7 | יאמר | 83.95 | **87.26** | 87.82 | 84.63 | .859 | 78.46 | 81.74 | 82.51 | 79.34 | .805 | **87.60** | 86.81 | **87.95** | **88.68** | .878 |
| 8 | מסכת | **97.06** | 95.79 | 92.44 | **94.66** | .950 | 90.84 | 87.18 | 78.04 | 83.82 | .849 | 96.99 | **96.45** | **93.53** | 94.49 | .954 |
| 9 | עם | 95.13 | **97.80** | 88.37 | 76.98 | **.894** | 92.61 | 97.32 | 83.89 | 64.27 | .838 | **95.30** | 97.36 | 86.50 | **77.90** | .891 |
| 10 | פניה | 90.25 | 87.32 | 96.90 | 97.68 | .930 | 81.24 | 76.75 | 94.35 | 95.63 | .870 | **93.76** | **87.97** | **97.08** | **98.56** | .943 |
| 11 | פרשו | **96.76** | 96.18 | 89.84 | 91.30 | .935 | 92.39 | 95.36 | 86.25 | 78.74 | .881 | 96.26 | **98.28** | **95.06** | 89.68 | .948 |
| 12 | שלישית | 94.44 | **95.90** | **96.15** | 94.77 | .953 | 90.52 | 88.37 | 89.47 | 91.43 | .899 | **96.16** | 94.35 | 94.86 | **96.51** | .955 |
| 13 | אחר | **97.56** | 97 72 | 95 47 | **95.17** | .965 | 95 02 | 96.96 | 93 72 | 89.94 | 939 | 97 39 | **98.44** | **96.84** | 94 77 | .969 |
| 14 | בניה | **93.54** | 9.85 | 97.85 | 98.51 | .952 | 82.57 | 70.68 | 93.28 | 96.47 | .855 | 92.68 | 90.17 | **97.69** | **98.31** | .947 |
| 15 | חזרה | 92.98 | 90 10 | 92 38 | 94.63 | .925 | 84 88 | 83.92 | 87.43 | 88.21 | 861 | **93.40** | **91.96** | **93.73** | **94.88** | .935 |
| 16 | ידע | 94.05 | 93.91 | 97.37 | 97.43 | .957 | 87.17 | 87.82 | 94.72 | 94.41 | .910 | **94.40** | **95.25** | **97.94** | **97.56** | .963 |
| 17 | כשר | 96.67 | 94.99 | 93.86 | 95.90 | .953 | 90 94 | 89 17 | 86 75 | 88.86 | 889 | **96.93** | **96.63** | **95.79** | **96.16** | .964 |
| 18 | כתב | **98.56** | **99.10** | **97.26** | 95.69 | **.976** | 94.65 | 96.57 | 89.11 | 83.71 | .910 | 98.51 | 98.65 | 95.95 | **95.56** | .972 |
| 19 | מבין | **96.73** | 96 31 | 98.64 | **98.80** | .976 | 96 42 | 96.57 | 97.63 | 98.72 | 966 | 96 12 | **96.74** | **98.80** | 98.56 | .976 |
| 20 | ספריה | 89.05 | 90.88 | 96.47 | 95.71 | .930 | 80.27 | 77.79 | 91.57 | 92.66 | .856 | **90.67** | **91.47** | **96.71** | **96.38** | .938 |
| 21 | עמנו | 89.58 | **88.87** | **94.67** | 95.03 | .920 | 86.07 | 83.50 | 92.18 | 93.51 | .888 | **91.48** | 87.48 | 94.11 | **96.08** | .923 |

Table 4: Full breakdown of the performance of our specialized classifiers when trained with short contexts (concatenation of encodings of the three word neighbors before and after the homograph). We display results for each of our three methods of encoding context words.

| | | Word2vec embeddings - BiLSTM | | | | | Morphological characteristics - BiLSTM | | | | | Composite Method - BiLSTM | | | | |
|---|---|---|---|---|---|---|---|---|---|---|---|---|---|---|---|---|
| | | Option 1 | | Option 2 | | | Option 1 | | Option 2 | | | Option 1 | | Option 2 | | |
| # | Word | Prec | Recall | Prec | Recall | Avg-F1 | Prec | Recall | Prec | Recall | Avg-F1 | Prec | Recall | Prec | Recall | Avg-F1 |
| 1 | את | 97.71 | **99.40** | **96.54** | 87.82 | .953 | 97.85 | 98.36 | 91.14 | 88.66 | .940 | **98.29** | 99.08 | 94.96 | **90.97** | **.958** |
| 2 | אתה | 95.62 | 96.72 | 96.88 | 95.83 | .963 | 92.35 | 94.16 | 94.41 | 92.67 | .934 | **95.65** | **97.61** | **97.71** | 95.83 | **.967** |
| 3 | אתכם | 94 16 | 95 22 | 90.20 | 88 14 | 919 | 88 29 | 89 55 | 78.42 | 76.17 | 831 | **95.51** | **96.54** | **92.90** | **90.90** | **.940** |
| 4 | אתם | **94.74** | 93.49 | 96.07 | 96.84 | .953 | 91.00 | 92.37 | 95.32 | 94.44 | .933 | 94.11 | **95.66** | **97.33** | 96.36 | **.959** |
| 5 | ברכת | 95 86 | **96.93** | **97.66** | 96.84 | **.968** | 78 33 | 76.95 | 82.81 | 83.92 | 805 | **96.09** | 95 91 | 96.91 | **97.05** | 965 |
| 6 | הרי | 98.95 | 98.60 | 93.75 | 95.24 | .966 | 96.98 | 98.15 | 91.13 | 86.17 | .931 | **99.00** | **98.75** | 94.39 | **95.46** | **.969** |
| 7 | יאמר | 87.10 | **91.32** | **91.63** | 87.54 | **.893** | 84 15 | 85.01 | 86.06 | 85.25 | 851 | 86 71 | 89 74 | 90.24 | 87 33 | .885 |
| 8 | מסכת | **98.40** | 97.57 | 95.59 | **97.07** | **.972** | 90.89 | 90.55 | 82.74 | 83.30 | .869 | 97.48 | **97.75** | **95.85** | 95.35 | .966 |
| 9 | עם | **96.37** | **97.92** | **89.66** | **83.06** | **.917** | 93 79 | 96.12 | 79.83 | 70.72 | 850 | 96 25 | 97 64 | 88 36 | 82 50 | 911 |
| 10 | פניה | 91.83 | 89.59 | 97.45 | 98.04 | .942 | 84.06 | 81.46 | 95.47 | 96.19 | .893 | **92.79** | 89.92 | **97.53** | **98.28** | **.946** |
| 11 | פרשו | **97.61** | 97 75 | 93.90 | **93.52** | 957 | 94 07 | 97 31 | 91 96 | 83.40 | 916 | 97.41 | **98.65** | **96.23** | 92 91 | **.963** |
| 12 | שלישית | **97.51** | 96.07 | **96.41** | **97.73** | **.969** | 92.31 | 91.48 | 92.18 | 92.95 | .922 | 96.86 | 96.07 | 96.39 | 97.12 | .966 |
| 13 | אחר | **98.21** | 98.64 | 97.28 | **96.43** | .976 | 94.64 | 96.80 | 93.36 | 89.14 | .935 | 97.90 | **98.96** | **97.89** | 95.80 | .976 |
| 14 | בניה | 92.93 | 95.76 | **98.99** | 98.27 | **.965** | 85.90 | 76.44 | 94.56 | 97.03 | .883 | **96.12** | 92.37 | 98.21 | **99.12** | .964 |
| 15 | חזרה | 95.19 | 93.81 | **95.18** | 96.26 | **.951** | 89.26 | 86.49 | 89.60 | 91.79 | .893 | **95.74** | 92.68 | 94.37 | **96.75** | .949 |
| 16 | ידע | 94.55 | 92.72 | 96.88 | 97.69 | .955 | 86.30 | 87.96 | 94.75 | 93.96 | .907 | **95.38** | **95.10** | **97.88** | **98.01** | **.966** |
| 17 | כשר | **98.75** | 96.63 | 95.89 | **98.46** | **.974** | 92.66 | 91.52 | 89.53 | 90.91 | .912 | 98.54 | 96.32 | 95.52 | 98.21 | .971 |
| 18 | כתב | 98.97 | **99.28** | **97.83** | 96.90 | .982 | 96.42 | 95.90 | 87.95 | 89.37 | .924 | **99.23** | 99.10 | 97.32 | **97.71** | **.983** |
| 19 | מבין | **97.23** | 95.44 | 98.33 | **99.00** | .975 | 95.95 | 95.22 | 98.24 | 98.52 | .970 | 96.77 | **97.50** | **99.08** | 98.80 | **.980** |
| 20 | ספריה | 90.38 | 93.97 | **97.65** | 96.16 | .945 | 82.73 | 81.03 | 92.77 | 93.50 | .875 | **92.15** | **93.24** | 97.39 | **96.95** | **.949** |
| 21 | עמנו | 89.75 | 87.08 | 93.88 | 95.22 | .915 | 82.45 | 83.10 | 91.85 | 91.50 | .872 | **90.71** | **89.26** | **94.88** | **95.61** | **.926** |

Table 5: Full breakdown of the performance of our specialized classifiers when trained with a bi LSTM of the full sentence context. We display results for each of our three methods of encoding context words.